\documentclass{article}

\usepackage{microtype}
\usepackage{graphicx}
\usepackage{subfigure}
\usepackage{booktabs} 
\usepackage{color}

\usepackage[accepted]{icml2018}

\usepackage{algorithmic}
\usepackage{algorithm}
\usepackage{amsmath}
\usepackage{comment}
\usepackage{amsmath}
\usepackage{amssymb}
\newtheorem{theorem}{Theorem}
\newtheorem{lemma}{Lemma}

\def\E{\mathbb{E}}
\def\V{\mathbb{V}{\bf ar}}
\def\C{\mathbb{C}{\bf ov}}
\DeclareMathOperator*{\argmin}{arg\,min}
\DeclareMathOperator*{\argmax}{arg\,max}
\def\tr{{\bf tr}}
\newcommand{\tagthisaux}{%
\refstepcounter{equation}%
{\textnormal{(\theequation)}}%
}
\newcommand{\tagthisline}{\`\tagthisaux}

\icmltitlerunning{Discovering and Removing Exogenous State Variables and Rewards for Reinforcement Learning}

\begin{document}
\twocolumn[
\icmltitle{Discovering and Removing Exogenous State Variables and Rewards for Reinforcement Learning}




\begin{icmlauthorlist}
\icmlauthor{Thomas Dietterich}{osu}
\icmlauthor{George Trimponias}{huawei}
\icmlauthor{Zhitang Chen}{huawei}
\end{icmlauthorlist}

\icmlaffiliation{osu}{School of EECS, Oregon State University,
  Corvallis, OR, USA}
\icmlaffiliation{huawei}{Huawei Noah's Ark Lab, Hong Kong}
\icmlcorrespondingauthor{Thomas Dietterich}{tgd@cs.orst.edu}

\icmlkeywords{Reinforcement Learning, Exogenous Variables, Variance Reduction}

\vskip 0.3in
]



\printAffiliationsAndNotice{}  

\begin{abstract}
Exogenous state variables and rewards can slow down reinforcement learning by injecting uncontrolled variation into the reward signal. We formalize exogenous state variables and rewards and identify conditions under which an MDP with exogenous state can be decomposed into an exogenous Markov Reward Process involving only the exogenous state+reward and an endogenous Markov Decision Process defined with respect to only the endogenous rewards. We also derive a variance-covariance condition under which Monte Carlo policy evaluation on the endogenous MDP is accelerated compared to using the full MDP. Similar speedups are likely to carry over to all RL algorithms. We develop two algorithms for discovering the exogenous variables and test them on several MDPs. Results show that the algorithms are practical and can significantly speed up reinforcement learning.
\end{abstract}

\section{Introduction} \label{sec:introduction}

In many practical settings, the actions of an agent have only a limited effect on the environment. For example, in a wireless cellular network, system performance is determined by a number of parameters that must be dynamically controlled to optimize performance. We can formulate this as a Markov Decision Process (MDP) in which the reward function is the negative of the number of users who are suffering from low bandwidth.  However, the reward is heavily influenced by exogenous factors such as the number, location, and behavior of the cellular network customers. Customer demand varies stochastically as a function of latent factors (news, special events, traffic accidents). In addition, atmospheric conditions can affect the capacity of each wireless channel.  This high degree of stochasticity can confuse reinforcement learning algorithms, because during exploration, the expected benefit of trying action $a$ in state $s$ is hard to determine. Many trials are required to average away the exogenous component of the reward so that the effect of the action can be measured. For temporal difference algorithms, such as Q Learning, the learning rate will need to be very small. For policy gradient algorithms, the number of Monte Carlo trials required to estimate the gradient will be very large (or equivalently, the step size will need to be very small). In this paper, we analyze this setting and develop algorithms for automatically detecting and removing the effects of exogenous state variables. This accelerates reinforcement learning (RL).

This paper begins by defining exogenous variables and rewards and presenting the exogenous-endogenous (Exo/Endo) MDP decomposition. We show that, under the assumption that the reward function decomposes additively into exogenous and endogenous components, the Bellman equation for the original MDP decomposes into two equations: one for the exogenous Markov reward process (Exo-MRP) and the other for the endogenous MDP (Endo-MDP). Importantly, every optimal policy for the Endo-MDP is an optimal policy for the full MDP. Next we study conditions under which solving the Endo-MDP is faster (in sample complexity) than solving the full MDP. To do this, we derive dynamic programming updates for the covariance between the $H$-horizon returns of the Exo-MRP and the Endo-MDP, which may be of independent interest. The third part of the paper presents an abstract algorithm for automatically identifying the Exo/Endo decomposition. We develop two approximations to this general algorithm. One is a global scheme that computes the entire decomposition at once; the second is a faster stepwise method that can scale to large problems. Finally, we present experimental results to illustrate cases where the Exo/Endo decomposition yields large speedups and other cases where it does not.  

\section{MDPs with Exogenous Variables and Rewards}
We study discrete time MDPs with stochastic rewards \cite{Puterman1994,Sutton1998}; the state and action spaces may be either discrete or continuous. Notation: state space $\cal{S}$, action space $\cal{A}$, reward distribution $R\!:\cal{S}\times \cal{A}\mapsto \cal{D}(\Re)$ (where $\cal{D(\Re)}$ is the space of probability densities over the real numbers), transition function $P\!: \cal{S} \times \cal{A} \mapsto \cal{D(S)}$ (where $\cal{D(S)}$ is the space of probability densities or distributions over $\cal{S}$), starting state $s_0$, and discount factor $\gamma \in (0,1)$. We assume that for all $(s,a)\in {\cal S}\times {\cal A}$, $R(s,a)$ has expected value $m(s,a)$ and finite variance $\sigma^2(s,a)$. 

Suppose the state space can be decomposed into two subspaces ${\cal X}$ and ${\cal E}$ according to ${\cal S} = {\cal E}\times {\cal X}$. We will say that the subspace ${\cal X}$ is \emph{exogenous} if the transition function can be decomposed as 
$P(e',x'|e,x,a)=P(x'|x)P(e'|x,e,a)$ $\forall e,e'\in {\cal E},x,x'\in {\cal X},a\in {\cal A},$ 
where $s'=(e',x')$ is the state that results from executing action $a$ in state $s = (e,x)$.  We will say that an MDP is an \emph{Exogenous State MDP} if its transition function can be decomposed in this way. We will say it is an \emph{Additively Decomposable Exogenous State MDP} if its reward function can be decomposed as the sum of two terms as follows. Let $R_{exo}\!:\!{\cal X} \mapsto \cal{D}(\Re)$ be the exogenous reward distribution and $R_{end}\!:\!{\cal E} \times {\cal X} \mapsto \cal{D}(\Re)$ be the endogenous reward distribution such that $r = r_{exo} + r_{end}$ where $r_{exo} \sim R_{exo}(\cdot|x)$ with mean $m_{exo}(x)$ and variance $\sigma^2_{exo}(x)<\infty$ and $r_{end} \sim R_{end}(\cdot|e,x,a)$ with mean $m_{end}(e,x,a)$ and variance $\sigma^2_{end}(e,x,a)<\infty$.

\begin{theorem}For any Additively Decomposable Exogenous State MDP with exogenous state space ${\cal X}$, the $H$-step finite-horizon Bellman optimality equation can be decomposed into two separate equations, one for a Markov Reward Process involving only ${\cal X}$ and $R_{exo}$ and the other for an MDP (the endo-MDP) involving only $R_{end}$:
\vspace{-0.1in}
\begin{tabbing}
xxxx\=xxxx\=xxxx\=\kill
$\displaystyle V(e,x;h) = V_{exo}(x;h) + V_{end}(e,x;h)$\tagthisline{}\label{eqn:sum}\\
$\displaystyle V_{exo}(x;h) = m_{exo}(x;h) \;+ $\tagthisline\label{eqn:exo}\\
\>$\displaystyle\gamma \E_{x'\sim P(x'|x)}[V_{exo}(x';h-1)]$\\ 
$\displaystyle V_{end}(e,x;h) = \max_{a} m_{end}(e,x,a)\; + $\tagthisline\label{eqn:end}\\
\> $\displaystyle \gamma \E_{x'\sim P(x'|x); e'\sim P(e'|e,x,a)}[V_{end}(e',x';h-1)].$ 
\end{tabbing}
\end{theorem}
\vspace{-0.1in}
\noindent{\bf Proof.} 
Proof by induction on the horizon $H$. Note that the expectations could be either sums (if $\cal{S}$ is discrete) or integrals (if $\cal{S}$ is continuous).

\noindent {\bf Base case:} $H=1$; we take one action and terminate. 
\vspace{-0.1in}
\[V(e,x;1) =  m_{exo}(x) + \max_a m_{end}(x,a).\]
The base case is established by setting $V_{exo}(x;1) = m_{exo}(x)$ and $V_{end}(e,x;1) = \max_a m_{end}(e,x,a)$. 

\noindent {\bf Recursive case:} $H=h$. 
\begin{tabbing}
xx\=xxxx\=xxxx\=\kill
$\displaystyle V(e,x;h) = m_{exo}(x) + \max_a \{m_{end}(e,x,a) \;+$ \\
\> $\displaystyle \gamma E_{x'\sim P(x'|x); e'\sim P(e'|e,x,a)}[V_{exo}(x';h-1) \;+$\\
\> \> $\displaystyle V_{end}(e',x';h-1)]\}$
\end{tabbing}
Distribute the expectation over the sum in brackets and simplify. We obtain
\vspace{-0.1in}
\begin{tabbing}
xx\=xxxx\=xxxx\=\kill
$\displaystyle V(e,x;h) = m_{exo}(x) + \gamma \E_{x'\sim P(x'|x)}[V_{exo}(x';h-1)] \;+ $\\
\>$\displaystyle \max_a \{m_{end}(e,x,a) \;+$\\
\>\>$\displaystyle \gamma \E_{x'\sim P(x'|x); e'\sim P(e'|e,x,a)}[V_{end}(e',x';h-1)]\}$
\end{tabbing}
\vspace{-0.1in}
The result is established by setting
\vspace{-0.1in}
\begin{tabbing}
xxxx\=xxxx\=xxxx\=\kill
$\displaystyle V_{exo}(x;h) = m_{exo}(x) + \gamma \E_{x'\sim P(x'|x)}[V_{exo}(x';h-1)]$\\
$\displaystyle V_{end}(e,x;h) = \max_{a} m_{end}(e,x,a) \;+$\\
\>$\displaystyle \gamma \E_{x'\sim P(x'|x); e'\sim P(e'|e,x,a)}[V_{end}(e',x';h-1)].$
\end{tabbing}
\vspace{-0.1in}
QED.

\begin{lemma}
Any optimal policy for the endo-MDP of Equation~\ref{eqn:end} is an optimal policy for the full exogenous state MDP.
\end{lemma}
\vspace{-0.1in}
\noindent{\bf Proof.} Because $V_{exo}(s;H)$ does not depend on the policy, the optimal policy can be computed simply by solving the endo-MDP. QED.  

We will refer to Equations~\ref{eqn:sum}, \ref{eqn:exo} and \ref{eqn:end} as the Exo/Endo Decomposition of the full MDP. 

In an unpublished manuscript, Bray (\citeyear{Bray2017}) proves a similar result. He also identifies conditions under which value iteration and policy iteration on the Endo-MDP can be accelerated by computing the eigenvector decomposition of the endogenous transition matrix. While such techniques are useful for MDP planning with a known transition matrix, we do not know how to exploit them in reinforcement learning where the MDP is unknown.

In other related work, McGregor et al. (\citeyear{McGregor2017}) show how to remove known exogenous state variables in order to accelerate an algorithm known as Model Free Monte Carlo \cite{Fonteneau2012}. Their experiments obtain substantial improvements in policy evaluation and reinforcement learning.

\section{Analysis of the Exo/Endo Decomposition}
Suppose we are given the decomposition of the state space into exogenous and endogenous subspaces. Under what conditions would reinforcement learning on the endogenous MDP be more efficient than on the original MDP? To explore this question, let us consider the problem of estimating the value of a fixed policy $\pi$ in a given start state $s_0$ via Monte Carlo trials of length $H$. We will compare the sample complexity of estimating $V^\pi(s_0;H)$ on the full MDP to the sample complexity of estimating $V^\pi_{end}(s_0;H)$ on the Endogenous MDP.  Of course most RL algorithms must do more than simply estimate $V^\pi(s_0;H)$ for fixed $\pi$, but the difficulty of estimating $V^\pi(s_0;H)$ is closely related to the difficulty of fitting a value function approximator or estimating the gradient in a policy gradient method.

Define $B^\pi(s_0;H)$ to be a random variable for the $H$-step cumulative discounted return of starting in state $s_0$ and choosing actions according to $\pi$ for $H$ steps. 
To compute a Monte Carlo estimate of $V^\pi(s_0;H)$, we will generate $N$ realizations $b_1, \ldots, b_N$ of $B^\pi(s_0;H)$ by executing $N$ $H$-step trajectories in the MDP, each time starting in  $s_0$. 
\begin{theorem}\label{theorem:chebychev}
For any $\epsilon>0$ and any $0<\delta<1$, let $\hat{V}^\pi(s_0;H)=1/N \sum_{i=0}^N b_i$ be the Monte Carlo estimate of the expected $H$-step return of policy $\pi$ starting in state $s_0$. If 
\vspace{-0.05in}
\[N \geq \frac{\V[B^\pi(s_0;H)]}{\delta \epsilon^2}\]
\vspace{-0.05in}
then
\vspace{-0.05in}
\[P[|\hat{V}^\pi(s_0;H) - V^\pi(s_0;H)| > \epsilon] \leq \delta.\]
\end{theorem}
\vspace{-0.05in}
{\bf Proof.} This is a simple application of the Chebychev Inequality, 
\[P(|X - \E[X]| > \epsilon) \leq \frac{\V[X]}{\epsilon^2},\]
with $X=\hat{V}^\pi(s_0;H)$. The variance of the mean of $N$ iid random variables is the variance of any single variable divided by $N$. Hence $\V[\hat{V}^\pi(s_0;H)] = \V[B^\pi(s_0;H)]/{N}$.
To obtain the result, plug this into the Chebychev inequality, set the rhs equal to $\delta$, and solve for $N$. QED.

Now let us consider the Exo/Endo decomposition of the MDP. Let $B^\pi_x(s_0;H)$ denote the $H$-step return of the exogenous MRP and $B^\pi_e(s_0;H)$ denote the return of the endogenous MDP. Then $B_x^\pi(s_0;H)+B_e^\pi (s_0;H)$ is a random variable denoting the cumulative $H$-horizon discounted return of the original, full MDP. Let $\V[B^\pi(s_0;H))]$ be the variance of $B^\pi(s_0;H)$ and $\C[B_x^\pi(s_0;H), B_e^\pi(s_0;H)]$ be the covariance between them.

\begin{theorem}\label{theorem:covariance-condition}
The Chebychev upper bound on the number of Monte Carlo trials required to estimate $V^\pi(s_0;H)$ using the endogenous MDP will be reduced compared to the full MDP iff $\V[B_x^\pi(s_0;H)] > -2\;\C[B_x^\pi(s_0;H),B_e^\pi(s_0;H)].$
\end{theorem}
{\bf Proof.} From Theorem \ref{theorem:chebychev}, we know that the sample size bound using the endogenous MDP will be less than the required sample size using the full MDP when $\V[B_e^\pi(s_0;H)] < \V[B_x^\pi(s_0;H) + B_e^\pi(s_0;H)]$. The variance of the sum of two random variables is
\begin{tabbing}
xxx\=xxx\=xxx\=xxx\=\kill
$\displaystyle \V[B_x^\pi(s_0;H) + B_e^\pi(s_0;H)] \;=$\\
\>$\displaystyle  \V[B_x^\pi(s_0;H)] + \V[B_e^\pi(s_0;H)] \;+$\\
\>\>$\displaystyle 2 \C[B_x^\pi(s_0;H), B_e^\pi(s_0;H)].$
\end{tabbing}
Hence, $\V[B_e^\pi(s_0;H)] < \V[B_x^\pi(s_0;H) + B_e^\pi(s_0;H)]$ iff
$\V[B_x^\pi(s_0;H)] > -2 \C[B_x^\pi(s_0;H), B_e^\pi(s_0;H)].$
QED.

To evaluate this covariance condition, we need to compute the variance and covariance of the $H$-step returns. We derive dynamic programming formulas for these.

\begin{theorem}\label{theorem:variance}
The variance of the $H$-step return $B^\pi(s;H)$ can be computed as the solution to the following dynamic program:
\begin{tabbing}
xxx\=xxx\=xxx\=xxx\=\kill
$\displaystyle V^\pi(s;0) := 0; \quad \V[B^\pi(s;0)] := 0$\tagthisline\label{eqn:var-base}\\
$\displaystyle V^\pi(s;h) := m(s,\pi(s)) \;+$\\
\> $\displaystyle \gamma \E_{s'\sim P(s'|s,\pi(s))}[V^\pi(s';h-1)]$\tagthisline\label{eqn:var-value}\\
$\displaystyle \V[B^\pi(s;h)] := \sigma^2(s,\pi(s)) - V^\pi(s;h)^2 \;+$\\
\> $\displaystyle \gamma^2\E_{s'\sim P(s'|s,\pi(s))}[\V[B^\pi(s';h-1)]] \;+$\tagthisline\label{eqn:var-var}\\
\> $\displaystyle \E_{s'\sim P(s'|s,\pi(s)}[m(s,\pi(s))+\gamma V^\pi(s';h-1)]^2$\\
Equations \ref{eqn:var-value} and \ref{eqn:var-var} apply for all $h>0$.
\end{tabbing}
\end{theorem}
{\bf Proof.} Sobel (\citeyear{Sobel1982}) analyzed the variance of infinite horizon discounted MDPs with deterministic rewards. We modify his proof to handle a fixed horizon and stochastic rewards. We proceed by induction on $h$. To simplify notation, we omit the dependence on $\pi$.

{\bf Base Case:} $H=0$. This is established by Equations \ref{eqn:var-base}. \\
{\bf Inductive Step:} $H=h$. Write 
\[B(s;h)=R(s)+\gamma B(s';h-1),\]
where the rhs  involves the three random variables $R(s)$, $s'$, and $B(s';h-1)$. To obtain Equation \ref{eqn:var-value}, compute the expected value
\vspace{-0.1in}
\begin{tabbing}
xxx\=xxx\=xxx\=xxx\=\kill
$\displaystyle V(s;h) = \E_{B(s;h)}[B(s;h)] \;=$\\
\> $\displaystyle \E_{s',R(s),B(s';h-1)}[R(s)+\gamma B(s';h-1)],$\\[0.05in]
and take each expectation in turn.
\end{tabbing}
To obtain the formula for the variance, write the standard formula for the variance:
\vspace{-0.1in}
\begin{tabbing}
xxx\=xxx\=xxx\=xxx\=\kill
$\displaystyle \V[B(s;h)] = \E_{B(s;h)}[B(s;h)^2] - \E_{B(s;h)}[B(s;h)]^2$
\end{tabbing}
\vspace{-0.1in}
Substitute $R(s)+\gamma B(s';h-1)$ in the first term and simplify the second term to obtain
\begin{tabbing}
xxx\=xxx\=xxx\=xxx\=\kill
$\displaystyle = \E_{s',R(s),B(s';h-1)}\left[\{R(s)+\gamma B(s';h-1)\}^2\right] - V(s;h)^2$\\
Expand the square in the first term:\\
$\displaystyle = \E_{s',R(s),B(s';h-1)}[R(s)^2 + 2R(s)\gamma B(s';h-1) \;+ $\\
\> $\displaystyle \gamma^2B(s';h-1)^2] - V(s;h)^2$\\
Distribute the two innermost expectations over the sum:\\
$\displaystyle = \E_{s'}[\E_{R(s)}[R(s)^2] + 2m(s)\gamma V(s';h-1) \;+ $\\
\> $\displaystyle \gamma^2\E_{B(s';h-1)}[B(s';h-1)^2]] - V(s;h)^2$
\end{tabbing}
\vspace{-0.1in}
Apply the definition of variance in reverse to terms 1 and 3 in brackets:
\vspace{-0.1in}
\begin{tabbing}
xxx\=xxx\=xxx\=xxx\=\kill
$\displaystyle = \E_{s'}[\V[R(s)] + m(s)^2 + 2m(s)\gamma V(s';h-1) \;+ $\\
\> $\displaystyle \gamma^2\V[B(s;h-1)] + \gamma^2V(s';h-1)^2] - V(s;h)^2$\\
Factor the quadratic involving terms 1, 3, and 4:\\
$\displaystyle = \E_{s'}[\sigma^2(s) + [m(s)+\gamma V(s';h-1)]^2 \;+$\\
\> $\displaystyle \gamma^2\V[B(s';h-1)]] - V(s;h)^2$ 
\end{tabbing}
\vspace{-0.1in}
Finally, distribute the expectation with respect to $s'$ to obtain Equation~\ref{eqn:var-var}. 
QED.

\begin{theorem}\label{theorem:covariance}
The covariance between the exogenous $H$-step return
$B_x^\pi(x;H)$ and the endogenous $H$-step return $B_e^\pi(e,x;H)$ can be computed via the following dynamic
program:
\begin{tabbing}
xxx\=xxx\=xxx\=xxx\=\kill
$\displaystyle \C[B_x^\pi(x;0),B_e^\pi(e,x;0)]:=0$\tagthisline\label{eqn:cov-base}\\
$\displaystyle \C[B_x^\pi(x;h),B_e^\pi(e,x;h)]:=$\\
\>$\displaystyle \gamma^2 \E_{e'\sim P(e'|e,x),x'\sim P(x'|x)}$\\
\>\> $\displaystyle \big(\C[B_x^\pi(x';h-1),B_e^\pi(e',x';h-1)] \;+ $\tagthisline\label{eqn:cov-cov}\\
\>\> $\displaystyle  [m_x(x)+\gamma V_x(x';h-1)]\;\times$\\
\>\>\>$ \displaystyle[m_e(e,x,\pi(e,x))+\gamma V_e^\pi(e',x';h-1)]\big)$\\
\> $\displaystyle - V_x^\pi(x;h)V_e^\pi(e,x;h).$\\[.05in]
Equations \ref{eqn:cov-base} and \ref{eqn:cov-cov} apply for all $h>0$.
\end{tabbing}
\end{theorem}
{\bf Proof.} By induction. The base case is established by Equation
\ref{eqn:cov-base}. For the inductive case, we begin with the formula
for non-centered covariance: 
\begin{tabbing}
xxx\=xxx\=xxx\=xxx\=\kill
$\displaystyle \C[B_x(x;h),B_e(e,x;h)]=$\\
\> $\displaystyle \E_{B_x(x;h),B_e(e,x;h)}[B_x(x;h)B_e(e,x;h)] \;-$\\
\>\>\>$\displaystyle V_x(x;h)V_e(e,x;h)$
\end{tabbing}
Replace $B_x(x;h)$ by $R_x(x)+\gamma B_x(x';h-1)$ and $B_e(e,x;h)$ by $R_e(e,x)+\gamma B_e(e',x';h-1)$ and replace the expectations wrt $B_x(x;h)$ and $B_e(e,x;h)$ by expectations wrt the six variables $\{e',x',R_x(x),R_e(e,x),B_x(x';h-1),B_e(e',x';h-1)\}$. We will use the following abbreviations for these variables: $s'=\{e',x'\}$, $r=\{R_e(x),R_e(e,x)\}$, and $B=\{B_x(x';h-1),B_e(e',x';h-1)\}$.
\begin{tabbing}
xxx\=xxx\=xxx\=xxx\=\kill
$\displaystyle \C[B_x(x;h),B_e(e,x;h)]= -V_x(x;h)V_e(e,x;h)\;+$\\
\> $\displaystyle \E_{s',r,B}[(R_x(x)+\gamma B_x(x';h-1))\;\times$\\
\>\> $\displaystyle (R_e(e,x)+\gamma B_e(e',x';h-1))]$
\end{tabbing}
\vspace{-0.1in}
We will focus on the expectation term. Multiply out the two terms and distribute the expectations wrt $r$ and $B$:
\vspace{-0.1in}
\begin{tabbing}
xxx\=xxx\=xxx\=xxx\=\kill
$\displaystyle \E_{s'}[m_x(x)m_e(e,x)+\gamma m_x(x)V_e(e',x';h-1)\;+$\\
\>$\displaystyle \gamma m_e(e,x)V_x(x';h-1) \;+$\\
\>$\displaystyle  \gamma^2 \E_B[B_x(x';h-1)B_e(e',x';h-1)]]$
\end{tabbing}
\vspace{-0.1in}
Apply the non-centered covariance formula ``in reverse'' to term 4.
\vspace{-0.1in}
\begin{tabbing}
xxx\=xxx\=xxx\=xxx\=\kill
$\displaystyle \E_{s'}[m_x(x)m_e(e,x)+\gamma m_x(x)V_e(e',x';h-1)\;+$\\
\>$\displaystyle \gamma m_e(e,x)V_x(x';h-1) \;+$\\
\>$\displaystyle  \gamma^2V_x(x';h-1)V_e(e',x';h-1) \;+$\\
\>$\displaystyle \gamma^2 \C[B_x(x';h-1)B_e(e',x';h-1)]]$\\
Distribute expectation with respect to $s'$:\\
$\displaystyle m_x(x)m_e(e,x)+\gamma m_x(x)\E_{s'}[V_e(e',x';h-1)] \;+$\\
\>$\displaystyle \gamma m_e(e,x)\E_{s'}[V_x(x';h-1)] \;+$\\ \>$\displaystyle\gamma^2\E_{s'}[V_x(x';h-1)]\E_{s'}[V_e(e',x';h-1)] \;+$\\
\>$\displaystyle \gamma^2 \C[B_x(x';h-1)B_e(e',x';h-1)]]$
\end{tabbing}

Obtain Equation \ref{eqn:cov-cov} by factoring the first four terms, writing the expectations explicitly, and including the $-V_x(x;h)V_e(e,x;h)$ term. QED.

To gain some intuition for this theorem, examine the three terms on the right-hand side of Equation \ref{eqn:cov-cov}. The first is the ``recursive'' covariance for $h-1$. The second is the one-step non-centered covariance, which is the expected value of the product of the backed-up values for $V_e^\pi$ and $V_x^\pi$. The third term is the product of $V_e^\pi$ and $V_x^\pi$ for the current state, which re-centers the covariance. 

Theorems \ref{theorem:variance} and \ref{theorem:covariance} allow us to check the covariance condition of Theorem \ref{theorem:covariance-condition} in every state, including the start state $s_0$, so that we can decide whether to solve the original MDP or the endo-MDP.  Some special cases are  easy to verify. For example, if the mean exogenous reward $m_x(s)=0$, for all states, then the covariance condition reduces to $\sigma^2_x(s) > 0$.

\section{Algorithms for Decomposing an MDP into Exogenous and Endogenous Components}

In some applications, it is easy to specify the exogenous state variables, but in others, we must discover them from training data. Suppose we are given a database of $\{(s_i,a_i,r_i,s'_i)\}_{i=1}^N$ sample transitions obtained by executing one or more exploration policies in the full MDP. Assume each $s_i,s'_i\in \cal{S}$ is a $d$-dimensional real-valued vector and $a_i \in \cal{A}$ is a $c$-dimensional real-valued vector (possibly a one-hot encoding of $c$ discrete actions). In the following, we center $s_i$ and $s'_i$ by subtracting off the sample mean.

We seek to learn three functions $F_{exo}$, $F_{end}$, and $G$ parameterized by $w_{x}$, $w_{e}$, and $w_G$, such that $x = F_{exo}(s; w_{x})$ is the exogenous state, $e = F_{end}(s; w_{e})$ is the endogenous state, and $s = G(F_{exo}(s),F_{exo}(s);w_G)$ recovers the original state from the exogenous and endogenous parts. We want to capture as much exogenous state as possible subject to the constraint that we (1) satisfy the conditional independence relationship $P(s'|s,a) = P(x'|x) P(e',x'|e,x,a)$, and (2) we can recover the original state from the exogenous and endogenous parts. We formulate the decomposition problem as the following abstract optimization problem:
\begin{equation}\label{opt:1}
\begin{split}
&\argmax_{w_x, w_e, w_G} \E[\| F_{exo}(s'; w_x)\|] \\
&\mbox{subject to}\\
&\hat{I}(F_{exo}(s';w_x); [F_{end}(s; w_e), a] | F_{exo}(s;w_x)) < \epsilon\\
& \E[\|G(F_{exo}(s'; w_x), F_{end}(s'; w_e); w_G) - s'\|] < \epsilon'
\end{split}
\end{equation}
We treat the $(s_i,a_i,r_i,s'_i)$ observations as samples from the random variables $s,a,r,s'$ and the expectations are estimated from these samples. The objective is to maximize the expected ``size'' of $F_{exo}$; below we will instantiate this abstract norm. In the first constraint, $\hat{I}(\cdot;\cdot|\cdot)$ denotes the estimated conditional mutual information. Ideally, it should be 0, which implies that $P(x'|x,e,a)=P(x'|x)$. We only require it to be smaller than $\epsilon$. The second constraint encodes the requirement that the average reconstruction error of the state should be small. We make the usual assumption that $x'$ and $s'$ are independent conditioned in $x$, $s$, and $a$.

We now instantiate this abstract schema by making specific choices for the norm, $F_{exo}$, $F_{end}$, and $G$. Instantiate the norm as the squared $L_2$ norm, so the objective is to maximize the variance of the exogenous state. Define $F_{exo}$ as a linear projection from the full $d$-dimensional state space to a smaller $d_x$-dimensional space, defined by the matrix $W_x$. The projected version of state $s$ is $W_x^{\top}s$. Define the endogenous state $e$ to contain the components of $s$ not contained within $x$. From linear decomposition theory \cite{Jolliffe}, we know that for a fixed $W_x$ consisting of \textit{orthonormal} components ($W_x^{\top}W_x=\mathbb{I}_{d_x}$), the exogenous state using all $d$ dimensions is $x=F_{exo}(s)=W_xW_x^{\top}s$, and the endogenous state $e=F_{end}(s)=s-x=s-W_xW_x^{\top}s$. Under this approach, $G(x,e)=F_{exo}(s)+F_{end}(s)=x+e=s$, so the reconstruction error is 0, and the second constraint in (\ref{opt:1}) is trivially satisfied. The endo-exo optimization problem becomes
\begin{equation}\label{opt:2}
\begin{split}
&\max_{0\leq d_x\leq d, W_x\in \mathbb{R}^{d\times d_x}}\E[\|W_x^{\top}s'\|^2_2] \\
&\mbox{subject to } W_x^{\top}W_x=\mathbb{I}_{d_x}\\
&\hat{I}(W_x^\top s'; [s-W_xW_x^{\top}s, a] | W_x^{\top}s)<\epsilon.
\end{split}
\end{equation}
Formulation \eqref{opt:2} involves simultaneous optimization over the unknown dimensionality $d_x$ and the projection matrix $W_x$. It is hard to solve, because of the conditional mutual information constraint. To tackle this, we approximate $\hat{I}(X;Y|Z)$ by the partial correlation coefficient $PCC(X,Y|Z)$, defined as the Frobenius norm of the normalized partial covariance matrix $V(X,Y,Z)$ \cite{baba2004partial,fukumizu2008kernel,fukumizu2004dimensionality}:
\[PCC(X;Y|Z)= \tr(V^\top(X,Y,Z)V(X,Y,Z))\]
where
\[V(X,Y,Z) = \Sigma_{XX}^{-1/2}(\Sigma_{XY} - \Sigma_{XZ}\Sigma_{ZZ}^{-1}\Sigma_{ZY})\Sigma_{YY}^{-1/2},\]
and ${\tr}$ is the trace.
If the set of random variables $(X,Y,Z)$ follow a multivariate Gaussian distribution, then the partial correlation coefficient of a pair of variables given all the other variables is equal to the conditional correlation coefficient \cite{baba2004partial}. With this change, we can express our optimization problem in matrix form as follows. Arrange the data into matrices $S,S',X,X',E,E'$, each with $n$ rows and $d$ columns, where the $i$th row refers to the $i$th instance. Let $A$ be the action matrix with $n$ rows and $c$ columns:
\begin{equation}\label{opt:3}
\begin{split}
&\max_{0\leq d_x\leq d, W_x\in \mathbb{R}^{d\times d_x}}\tr(W_x^{\top}S'^{\top}S'W) \\
&\mbox{subject to } W_x^{\top}W_x=\mathbb{I}_{d_x}\\
&PCC(S'W_x; [S-SW_xW_x^{\top}, A] | SW_x)<\epsilon.
\end{split}
\end{equation}

\subsection{Global Algorithm}
Formulation \eqref{opt:3} is challenging to solve directly because of the second constraint, so we seek an approximate solution. We simplify \eqref{opt:3} as
\begin{equation}\label{opt:4}
\begin{split}
&\min_{W_x\in \mathbb{R}^{d\times d_x}} PCC(S'W_x; [S-SW_xW_x^{\top}, A] | SW_x)\\
&\mbox{subject to } W_x^{\top}W_x=\mathbb{I}_{d_x}
\end{split}
\end{equation}
This formulation assumes a \textit{fixed} exogenous dimensionality $d_x$ and computes the projection with the minimal PCC. The orthonormality condition constrains $W_x$ to lie on a Stiefel manifold \cite{Stiefel1935}. Several optimization algorithms exist for optimizing on Stiefel manifolds \cite{Jiang2015,Absil2007,Edelman1999}. We employ the algorithms implemented in the  Manopt package \cite{Boumal2014}.

For a fixed dimensionality $d_x$ of the exogenous state, we can solve \eqref{opt:4} to minimize the PCC. Given that our true objective is \eqref{opt:3} and the optimal value of $d_x$ is unknown, we must try all values $d_x=0,\ldots,d$, and pick the $d_x$ that achieves the maximal variance, provided that the PCC from optimization problem \eqref{opt:4} is less than $\epsilon$. 
This is costly since it requires solving $O(d)$ manifold optimization problems. We can speed this up somewhat by iterating $d_x$ from $d$ down to 1 and stopping with the first projection $W_x$ that satisfies the PCC constraint. If no such projection exists, then the exogenous projection is empty, and the state is fully endogenous. The justification for this early-stopping approach is that the exogenous decomposition that maximizes $d_x$ will contain any component that appears in exogenous decompositions of lower dimensionality. Hence, the $W_x$ that attains maximal variance is achieved by the largest $d_x$ satisfying the PCC constraint. Algorithm \ref{alg:Global} describes this procedure. Unfortunately, this scheme must still solve $O(d)$ optimization problems in the worst case. We now introduce an efficient stepwise algorithm.
\begin{algorithm}
\begin{algorithmic}[1]
\STATE {\bf Inputs:} A database of transitions $\{(s_i,a_i,r_i,s'_i)\}_{i=1}^N$
\STATE {\bf Output:} The exogenous state projection matrix $W_x$ 
\FOR {$d_x=d$ \textbf{down to} $1$}
\STATE Solve the following optimization problem:
\vspace{-0.1in}
\begin{align*}
W_x := \quad\quad&\\
\argmin_{W\in \mathbb{R}^{d\times  d_x}}&PCC(S'W; [S-SWW^{\top}, A] | SW)\\
\textrm{subject to } &W^{\top}W=\mathbb{I}_{d_x}
\end{align*}
\vspace{-0.25in}
\STATE $pcc \leftarrow PCC(S'W_x; [S-SW_xW_x^{\top}, A] | SW_x)$
\IF {$pcc < \epsilon$}
\RETURN $W_x$
\ENDIF
\ENDFOR
\RETURN empty projection $[~]$
\end{algorithmic}
\caption{Global Exo/Endo State Decomposition}
\label{alg:Global}
\end{algorithm}

\subsection{Stepwise Algorithm}
Our stepwise algorithm extracts the components (vectors) of the exogenous projection matrix $W_x$ one at a time by solving a sequence of small optimization problems.  Suppose we have already identified the first $k-1$ components of $W_x$, $w_1, \ldots, w_{k-1}$ and we seek to identify $w_k$. These $k-1$ components define $k-1$ exogenous state variables $x_1, x_2, \ldots, x_{k-1}$. Recall that the original objective $PCC(X' ;[E,A]|X)$ seeks to uncover the conditional independence $P(x'|x,e,a) = P(x'|x)$. This requires us to know $x$ and $e$, whereas we only know a portion of $x$, and we therefore do not know $e$ at all. To circumvent this problem, we make two approximations. First, we eliminate $e$ from the conditional independence. Second, we assume that $P(X'|X)$ has a lower-triangular form, so that $P(x'_k| x,a) = P(x'_k | x_1, \ldots, x_{k-1}, x_k,a)$. 
This yields the simpler objective $PCC(X'_k; A|X_1, \ldots, X_k)$. 

What damage is done by eliminating $E$?  Components with low values for $PCC(X';[E,A]|X)$ will also have low values for $PCC(X';A|X)$, so this objective will not miss true exogenous components. On the other hand, it may find components that are not exogenous, because they depend on $E$. To address this, after discovering a new component we add it to $W_x$, only if it satisfies the original PCC constraint conditioned on both $E$ and $A$. What damage is done by assuming a triangular dependence structure?  We do not have a mathematical characterization of this case, so we will assess it experimentally.
\begin{algorithm}[b!]
\begin{algorithmic}[1]
\STATE {\bf Inputs:} A database of transitions $\{(s_i,a_i,r_i,s'_i)\}_{i=1}^N$
\STATE {\bf Output:} The exogenous state projection matrix $W_x$ 
\STATE Initialize projection matrix $W_x\leftarrow[~]$, $C_x\leftarrow[~]$, $k\leftarrow 0$
\REPEAT
\STATE $N\leftarrow \textrm{orthonormal basis for null space of }C_x$
\STATE Solve the following optimization problem:
\vspace{-0.1in}
\begin{align*}
\hat{w} := \quad\quad\quad&\\
\argmin_{w\in \mathbb{R}^{1\times(d-k)}}&PCC(S'[W_x,Nw^{\top}];A|S[W_x, Nw^{\top}])\\
\textrm{subject to } &w^{\top}w=1
\end{align*}
\vspace{-0.3in}
\STATE $w_{k+1} \leftarrow N\hat{w}^{\top}$ 
\STATE $C_x \leftarrow C_x \cup \{w_{k+1}\}$
\STATE $E=S-S[W_x,w_{k+1}][W_x,w_{k+1}]^{\top}$
\STATE $pcc \leftarrow PCC(S'[W_x,w_{k+1}];[E,A]|S[W_x, w_{k+1}])$
\IF {$pcc < \epsilon$}
\STATE $W_x \leftarrow W_x \cup \{w_{k+1}\}$
\ENDIF
\STATE $k\leftarrow k+1$
\UNTIL $k=d$
\RETURN $W_x$
\end{algorithmic}
\caption{Stepwise Exo/Endo State Decomposition}
\label{alg:Stepwise}
\end{algorithm}

To finish the derivation of the stepwise algorithm, we must ensure that each new component is orthogonal to the previously-discovered components.  Consider the matrix $C_x=[w_1, w_2, \dots, w_k]$. To ensure that the new component $w_{k+1}$ will be orthogonal to the $k$ previously-discovered vectors, we restrict $w_k$ to lie in the null space of $C_x$. We compute an orthonormal basis $N$ for this null space.  The matrix $\overline{C}=[C_x\textrm{ }N]$ is then an orthonormal basis for $\mathbb{R}^{d}$. Since we want $w_{k+1}$ to be orthogonal to the components in $C_x$, it must have the form $[C_x\textrm{ }N]\cdot[\mathbf{0}^{1\times k}\textrm{ }w]^{\top}=\overline{C}\cdot[\mathbf{0}^{1\times k}\textrm{ }w]^{\top}=Nw^{\top}$, where $w\in \mathbb{R}^{1\times (d-k)}$. The unit norm constraint $\|\overline{C}\cdot[\mathbf{0}^{1\times k}\textrm{ }w]^{\top}\|_2=1$ is satisfied if and only if $\|w\|_2=1$.

Algorithm \ref{alg:Stepwise} incorporates the aforementioned observations. Line 5 computes an orthonormal basis for the null space $N$ of all components $C_x$ that have been discovered so far. Lines 6-8 compute the component in $N$ that minimizes the PCC and add it to $C_x$. Line 9 computes the endogenous space $E$ assuming the new component was added to the current exogenous projection $W_x$. In Lines 10-13, if the PCC, conditioned on both the endogenous state and the action, is lower than $\epsilon$, then we add the newly discovered component to $W_x$. The algorithm terminates when the entire state space of dimensionality $d$ has been decomposed.

As in the global algorithm, Line 6 involves a manifold optimization problem; however, the Stiefel manifold corresponding to the constraint $w^{\top}w=1$ is just the unit sphere, which has a simpler form than the general manifold in formulation \eqref{opt:4} in the Global algorithm \ref{alg:Global}. Note that the variance of the exogenous state can only increase as we add components to $W_x$. The stepwise scheme has the added benefit that it can terminate early, for instance once it has discovered a sufficient number of components or once it has exceeded a certain time threshold.

\section{Experimental Studies}
We report three experiments. (Several additional experiments are described in the Supplementary Materials.) In each experiment, we compare Q Learning \cite{watkins1989learning} on the full MDP (``Full MDP'') to Q Learning on the decomposed MDPs discovered by the Global (``Endo MDP Global'') and Stepwise (``Endo MDP Stepwise'') algorithms. Where possible, we also compare the performance of Q Learning on the true endogenous MDP (``Endo MDP Oracle''). The Q function is represented as a neural network with a single hidden layer of $20$ tanh units and a linear output layer, except for Problem 1 where 2 layers of $40$ units each are used. Q-learning updates are implemented with stochastic gradient descent. Exploration is achieved via Boltzmann exploration with temperature parameter $\beta$. Given the current Q values, the action $a_t$ is selected according to 
\begin{equation}
a_t \sim \pi(a|s_t) = \frac{\exp(Q(s_t,a)/\beta)}{\sum_i \exp(Q(s_t,a_i)/\beta)}.
\end{equation}

In each experiment, all Q learners observe the entire current state $s_t$, but the full Q learner is trained on the full reward, while the endogenous reward Q learners are trained on the (estimated) endogenous reward. All learners are initialized identically and employ the same random seed. For the first $L$ steps, the full reward is employed, and we collect a database of $(s,a,s',r)$ transitions.  We then apply Algorithm~\ref{alg:Global} and  Algorithm~\ref{alg:Stepwise} to this database to estimate $W_x$ and $W_e$. The algorithms then fit  a linear regression model $R_{exo}(W_x^\top s)$ to predict the reward $r$ as a function of the exogenous state $x=W_x^\top s$. The endogenous reward is defined as the residuals of this model: $R_{end}(s) = r - R_{exo}(x)$. The endogenous Q learner then employs this endogenous reward for steps $L+1$ onward. Each experiment is repeated $N$ times. A learning curve is created by plotting one value every $T$ steps, which consists of the mean of $N\times T$ immediate rewards, along with a 95\% confidence interval for that mean.

In all Q Learners, we set the discount factor to be $0.9$. The learning rates are set to $0.001$ for Problem 1, and $0.02$ for Problem 2, and $0.05$ for Problem 3. The temperature of Boltzmann exploration is set to $5.0$ for Problem 1 and $1.0$ for Problems 2 and 3. We employ steepest descent solving in Manopt. For the PCC constraint, $\epsilon$ is set to $0.05$.
\begin{figure}
\centering
\includegraphics[width=\columnwidth]{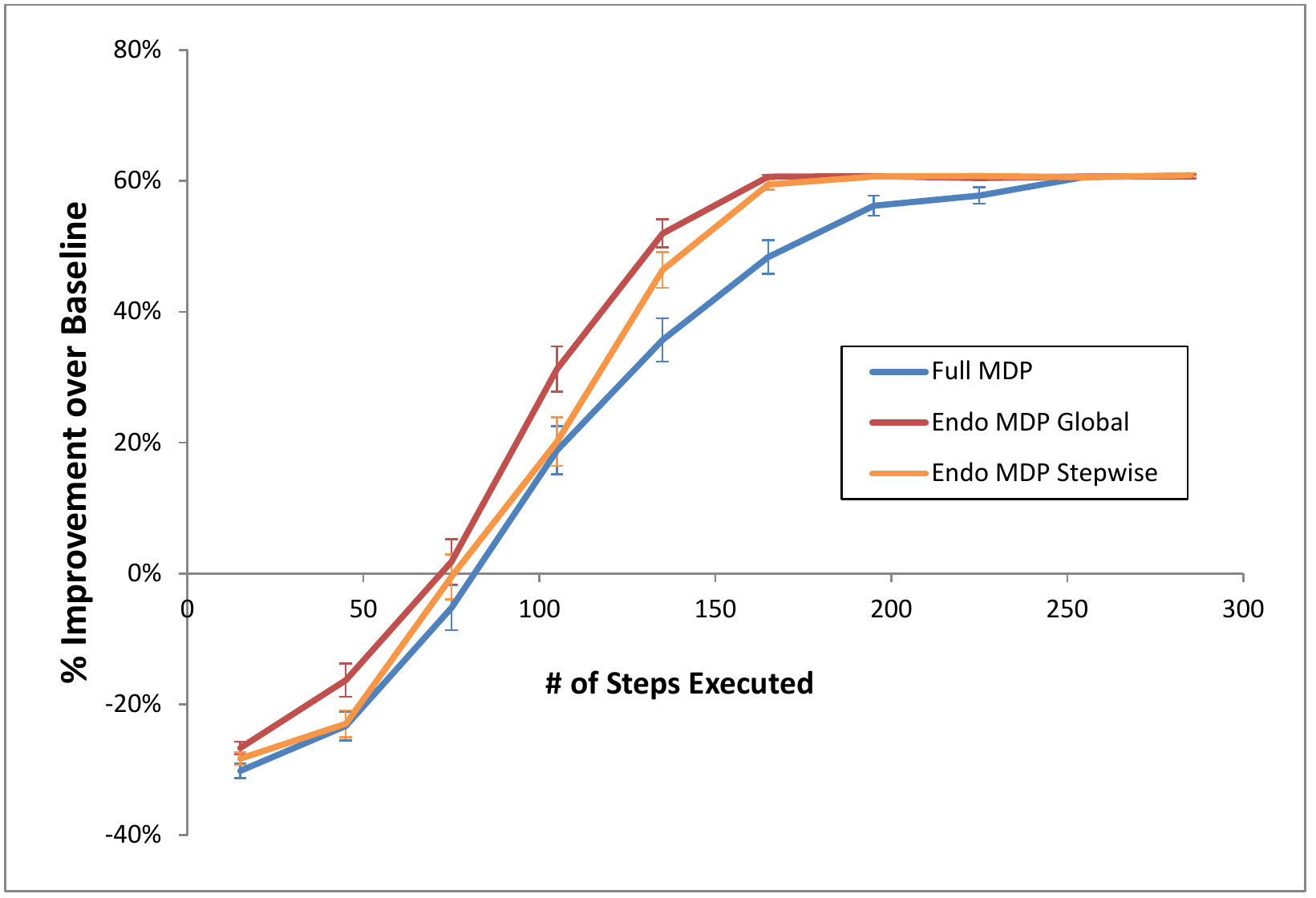}
\vspace*{-20pt}
\caption{Performance of Q Learning applied to the Full MDP and to the Endogenous MDP as estimated by the Global and Stepwise algorithms on the wireless network task ($N=20, T=30$). Performance is \% improvement over the performance of the configuration recommended by the manufacturer (shown as 0\%).}
\label{fig:wireless}
\end{figure}

\subsection{Problem 1: Wireless Network Parameter Configuration}
\label{Sec:wireless}

We begin by applying our developed algorithm to the wireless network problem described in Section~\ref{sec:introduction}. The parameter $a_t$ to be configured in the time period $[t,t+\Delta t]$ is the threshold of monitored signal power that determines when a user can switch to other frequency bands with stronger signals. The reward is $R_t = - P_t$, where  $P_t$ is the percentage of the users that suffer from low bandwidth during the time period. The state vector $S_t$ consists of five features: (1) the number of active users during the observation period; (2) the average number of users registered with the network during the period; (3) the channel quality index; (4) the ratio of small packets to total packets; and (5) the ratio of total bytes in small packets to total bytes from all packets. The discount factor is $\lambda = 0.95$. The overall reward depends on the number of active users, and this in turn is driven by exogenous factors that are not controlled by the policy. This observation motivated this research.

We obtained 5 days of hourly data collected from a cell network with 105 cells.  To create a simulator, we applied Model-Free Monte Carlo \cite{Fonteneau2012} to synthesize new trajectories by stitching together segments from the historical trajectories. We applied Q learning to the full MDP and to the endogenous MDP computed by both of our methods. Instead of measuring instantaneous performance, which is very noisy, we captured the Q function $Q_\tau$ at each iteration $\tau$ and ran independent evaluation trials starting in 10 selected states $\{s_1, \ldots, s_{10}\}$. Trial $i$ always starts in state $s_i$, applies the policy for $H=\frac{\log 0.01}{\log \lambda}$ steps, and returns the cumulative discounted reward $V_i^H(s_i) = \sum_{j=0}^H \lambda^jR_{ij}$. The estimated value of the policy $\pi_\tau$ (defined by $Q_\tau$) is $\hat{V}^{\pi_\tau}=1/10 \sum_{i=1}^{10} V_i^H(s_i)$. 

Figure \ref{fig:wireless} presents the results. We observe that the policies learned from the Global and Stepwise Endo MDPs converge in 165 steps. At that point, the full MDP has only achieved 73\% of the improvement achieved by the Endo MDPs. An additional 90 steps (54\%) are required for the full MDP to obtain the remaining 27\% points. 

\subsection{Problem 2: Two-dimensional Linear System with Anti-correlated Exogenous and Endogenous Rewards}
\label{Sec:bad_example}
The second problem is designed to test how the methods behave when the covariance condition is violated. Denote by $X_t$ the exogenous state and by $E_t$ the endogenous state at time $t$. The state transition function is 
\[X_{t+1} = 0.9 X_{t} + \epsilon_x;\;\; E_{t+1} = 0.9 E_{t} + A_{t} + 0.1 X_{t} + \epsilon_e,\]
where $\epsilon_x \sim \mathcal{N}(0,0.16)$ and $\epsilon_e \sim \mathcal{N}(0,0.04)$.

The observed state vector $S_t$ is a linear mixture of the hidden exogenous and endogenous states defined as $S_t = M [X_{t}, E_{t}]^{\top}$, where
$M=\begin{bmatrix}
0.4 & 0.6\\
0.7 & 0.3
\end{bmatrix}.$
The reward at time $t$ is $R_t = R_{x,t} + R_{e,t}$, where $R_{x,t}$ is the exogenous reward $R_{x,t} = \exp[-|X_t + 3|/5]$ and $R_{e,t}$ is the endogenous reward  $R_{e,t} = \exp[-|E_t - 3|/5]$. The optimal policy is to drive $E_{t}$ to $3$. Because the reward functions are anti-correlated, the condition of Theorem \ref{theorem:covariance-condition} is violated. The variance $\V[B_x] = 0.235$, which is less than $-2 \C[B_x,B_e] = 0.388$. 

\begin{figure}
\centering
\includegraphics[width=\columnwidth]{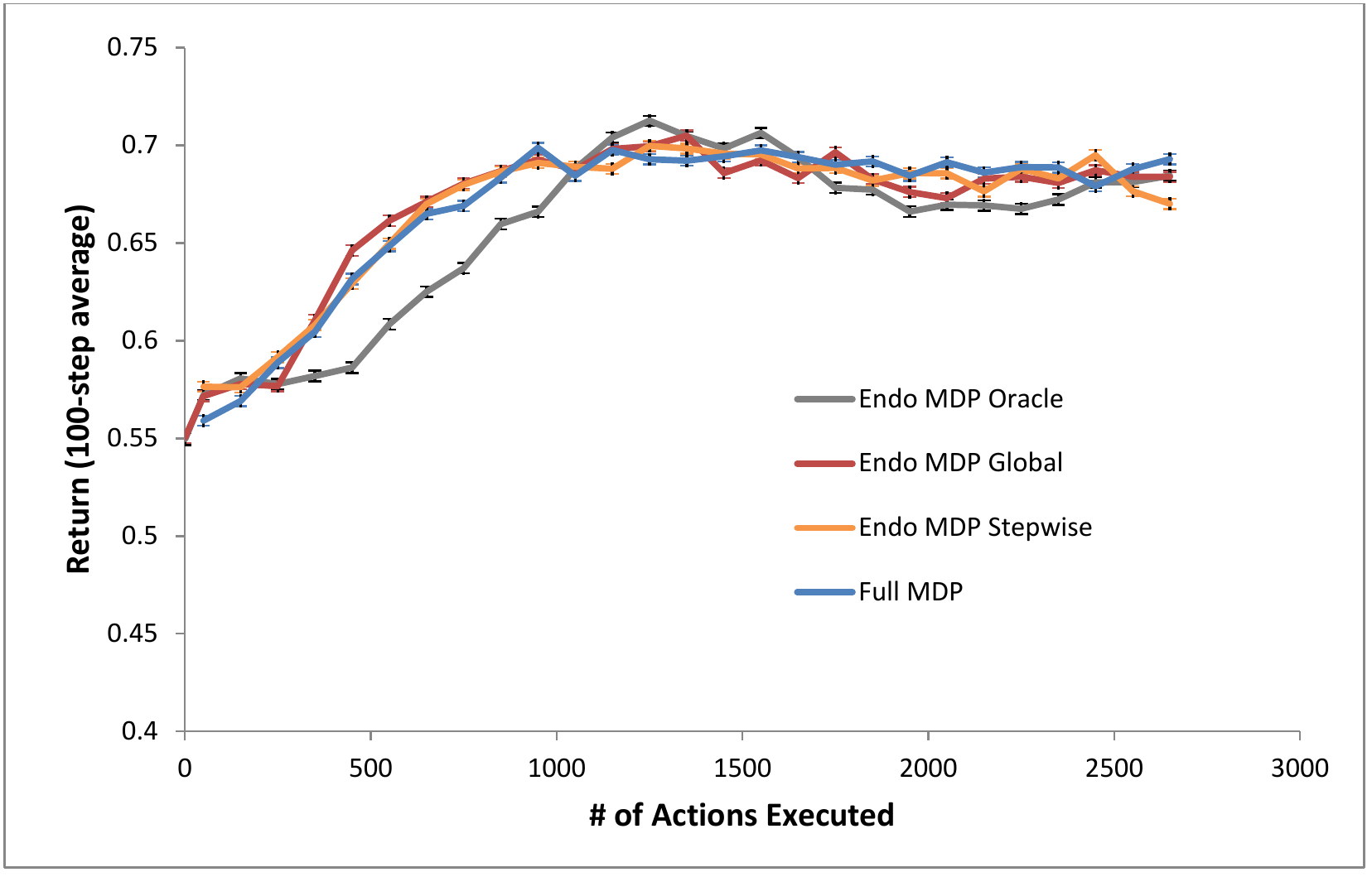}
\vspace*{-20pt}
\caption{Comparison of Q Learning applied to the Full MDP, to the Endogenous MDP, and to Endogenous Oracle on a 2-d linear MDP that fails the covariance condition of Theorem \ref{theorem:covariance-condition} ($N=200, T=100$). }
\label{fig:fail}
\end{figure}

Our theory predicts that the Endo-Q learners will learn more slowly than Q learning on the full MDP. The results (see Figure \ref{fig:fail}) provide partial support for this. As expected, the Oracle Endo MDP exhibits slower learning. But we were surprised to see that the Stepwise and Global methods discovered additional exogenous state that we did not realize was present. This reduced the negative correlation and allowed them to match the performance of Q learning on the full MDP. Hence, this experiment both confirms our theory and also shows that there can be ``hidden'' exogenous state that can be exploited to speed up learning.

\begin{figure}
\centering
\includegraphics[width=\columnwidth]{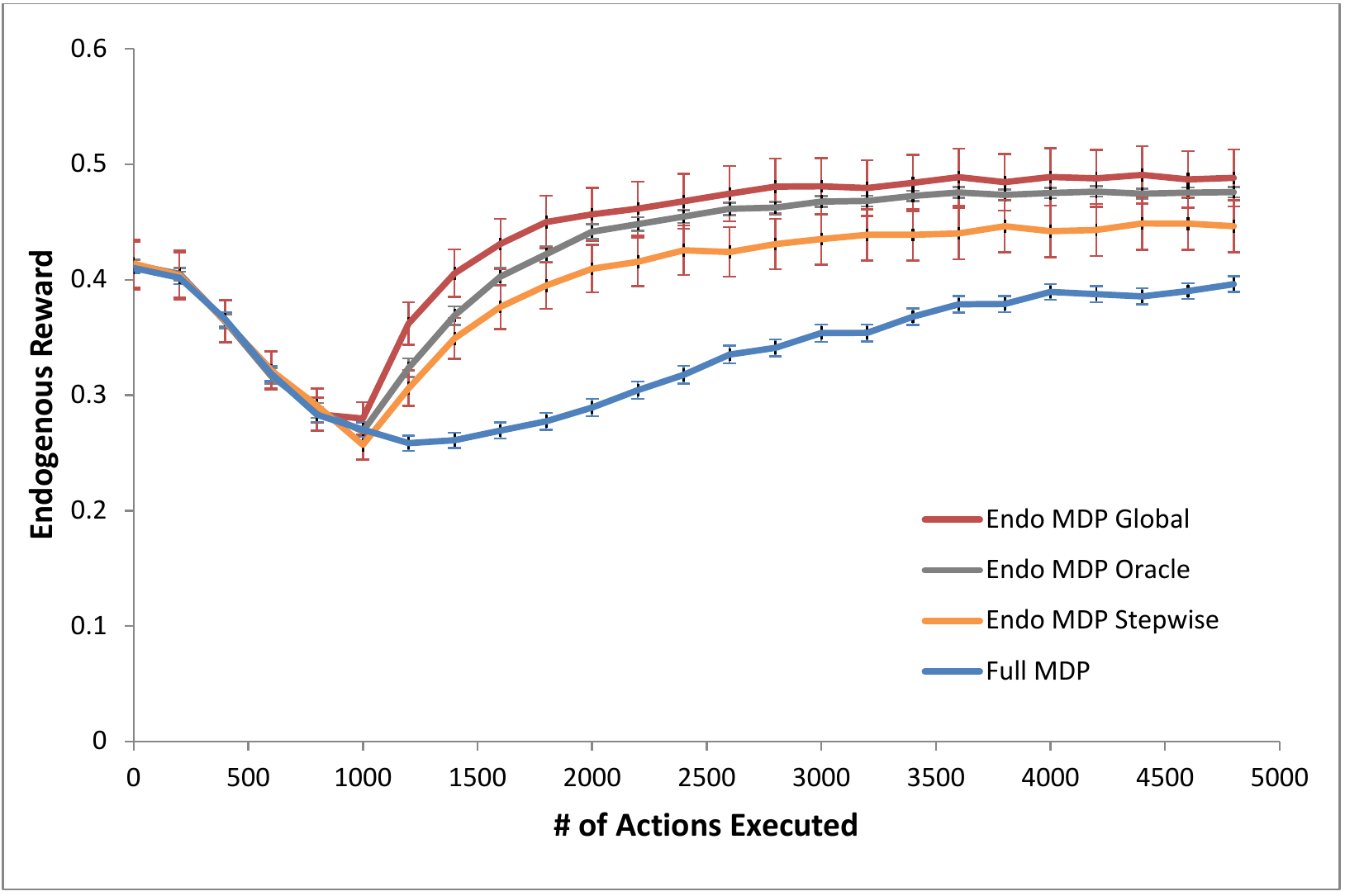}
\vspace*{-20pt}
\caption{Comparison of methods on the 30-dimensional linear system, $N=1500, T=1$.}
\label{fig:15x15}
\end{figure}
\subsection{Problem 3: High dimensional linear system}
\label{Sec:high-D}
The final experiment tests how well our algorithms can handle high-dimensional problems. We designed a 30-dimensional state MDP with 15 exogenous and 15 endogenous state variables. Let $X_t=[X_{1,t},\dots,X_{15,t}]^{\top}$ be the exogenous variables and $E_t=[E_{1,t},\dots,E_{15,t}]^{\top}$ be the endogenous variables. The state transition function is 
\[X_{t+1} = M_{x}\cdot X_{t} + \mathcal{E}_x;\;\; E_{t+1} = M_{e}\cdot \begin{bmatrix}E_{t}\\ X_{t}\\ A_{t}\end{bmatrix} + \mathcal{E}_e,\]
where $M_x\in\mathbb{R}^{15\times15}$ is the transition function for the exogenous MRP; $M_e\in\mathbb{R}^{15\times31}$ is the transition function for the endogenous MDP and involves all $E_{t}, X_{t},$ and $A_{t}$; 
$\mathcal{E}_x\in\mathbb{R}^{15\times1}$ is the exogenous noise, whose elements are distributed according to $\mathcal{N}(0,0.09)$; and  $\mathcal{E}_e\in\mathbb{R}^{15\times1}$ is the endogenous noise, whose elements are distributed according to $\mathcal{N}(0,0.04)$.
The observed state vector $S_t$ is a linear mixture of the hidden exogenous and endogenous states defined as $S_t = M\cdot\begin{bmatrix}E_{t}\\ X_{t}\end{bmatrix}$, where $M\in\mathbb{R}^{30\times30}$. The elements in  $M_x$, $M_e$, and $M$ are generated according to $\mathcal{N}(0,1)$ and then each row of each matrix is normalized to sum to 0.99 for stability. The starting state is the zero vector. The reward at time $t$ is $R_t = R_{x,t} + R_{e,t}$, where $R_{x,t}$ is the exogenous reward $R_{x,t} = -3\cdot\textrm{avg}(X_{t})$ and $R_{e,t}$ is the endogenous reward  $R_{e,t} = \exp[-|\textrm{avg}(E_t) - 1|]$, where $\textrm{avg}$ denotes the average over a vector's elements. 

Figure~\ref{fig:15x15} shows the true endogenous reward attained by four Q learning configurations. For the first 1000 steps, all methods are optimizing the full reward (not shown) which causes the endogenous reward to drop. Then the Endo-Exo decompositions are computed, and performance begins to improve. The Global and Stepwise methods substantially out-perform Q learning on the Full MDP. Endo Global is the quickest learner and even beats Endo Oracle, hence  it may have discovered additional exogenous state.

We must note that in Problems 2 and 3, Q learning needs to run much longer to discover the optimal policy (0.97 in Problem 2 and 0.92 in Problem 3). 

\section{Concluding Remarks}
This paper developed a theory of exogenous-state MDPs. It showed that if the reward decomposes additively, then the MDP can be decomposed into an exogenous Markov reward process and an endogenous MDP such that any optimal policy for the endogenous MDP is an optimal policy for the original MDP.  We derived a covariance criterion that elucidates the conditions under which solving the endogenous MDP can be expected to be faster than solving the full MDP. We then presented two practical algorithms (Global and Stepwise) for decomposing an MDP based on observed trajectories.  Experiments on synthetic problems and on a cellular network management problem confirmed that these algorithms can significantly speed up Q learning.  An important open question is how best to explore the MDP for purposes of learning the endo-exo decomposition.

\newpage
\section*{Acknowledgments}
Dietterich was supported by a gift from Huawei, Inc.  

\bibliography{exo}
\bibliographystyle{icml2018}

\newpage
\appendix
\section{Supplementary Materials}

We experimented with several additional test problems that did not fit into the main body of the paper. 

\subsection{Problem 1: Route Planning with Traffic}
\begin{figure}
\centering
	\includegraphics[width=\columnwidth]{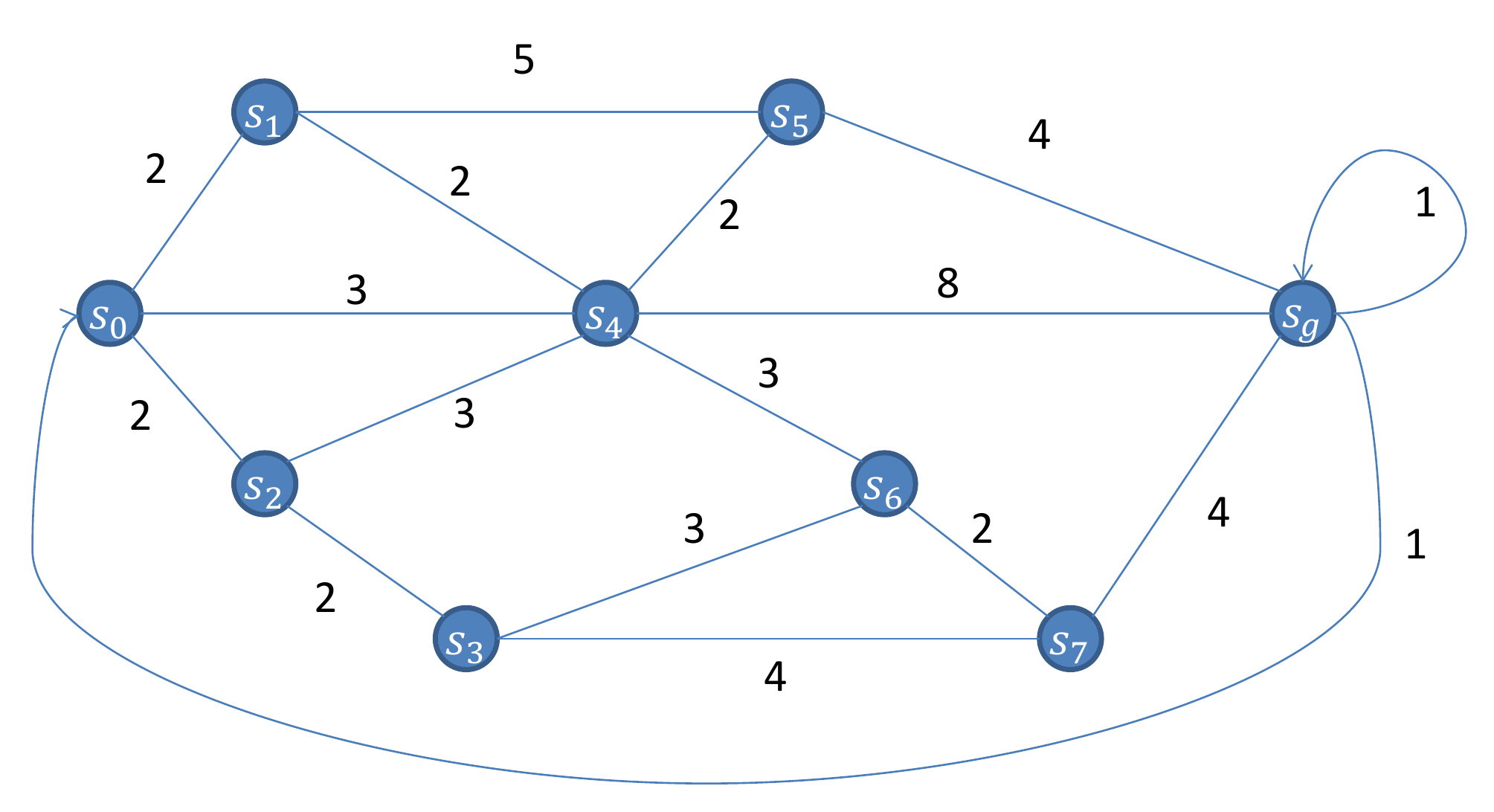}
	\caption{A traffic network. A car starts in $s_0$ and seeks to reach $s_g$ by the quickest path. The time to traverse each edge under ideal conditions is shown. Exogenous traffic can add delays to these times.}
	\label{fig:traffic-network}
\end{figure}

Consider training a self-driving car to minimize the time it takes to get to the office every morning. It is natural to reward the car for minimizing the travel time, but the primary factor determining the reward is the exogenous traffic of all the other drivers on the road. This is similar in many ways to the cellular network management problem. The purpose of this first problem is to see if our methods work on a simple version of this problem. Figure~\ref{fig:traffic-network} shows a road network MDP. The endogenous states are the nodes of the network. The exogenous state $X_t$ is the level of traffic in the network. It evolves according to the linear system $X_{t+1} = 0.9 X_t + \epsilon,$ where $\epsilon \sim \mathcal{N}(0, 1).$
The reward function is:
\[
r_t = \frac{1}{cost(s_t \rightarrow s_{t+1})} + X_t.\]
The actions at each node consist of choosing one of the outbound edges to traverse. To make the task easier, we restrict the set of actions to move only rightward (i.e., toward states with higher subscripts) except that $s_g$ can return to $s_0$. The cost of traversing an edge is shown by the weights in Figure ~\ref{fig:traffic-network}. For example, if the agent moves from $s_0$ to $s_4$, the $cost(s_0\rightarrow s_4)=3$.

The Q function is represented as a neural network with a 1-hot encoding of the 9 states plus a tenth input unit for $X_t$ and an eleventh input unit for $A_t$.  The PCC threshold $\epsilon = 0.05$.

\begin{figure}
\centering
	\includegraphics[width=\columnwidth]{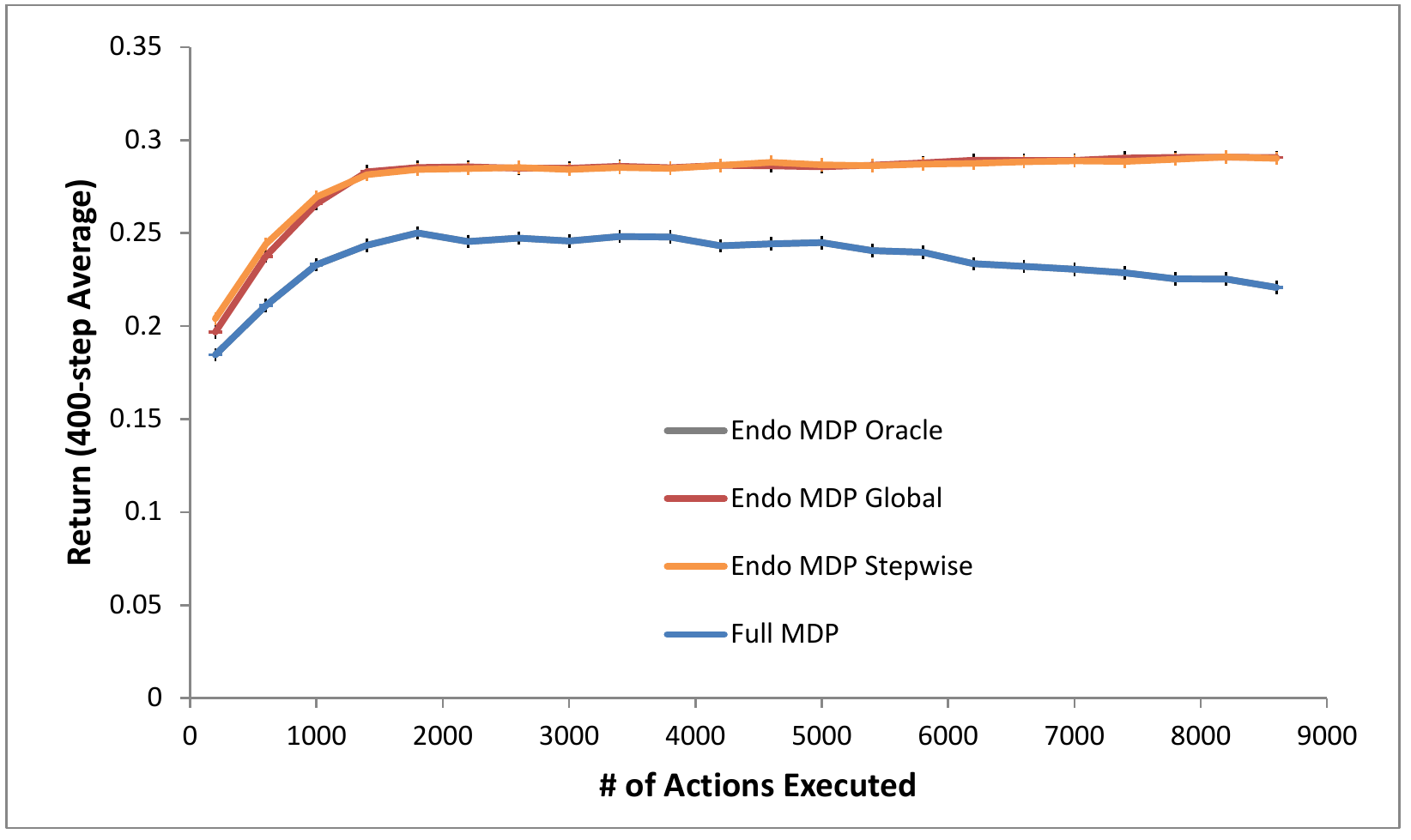}
	\caption{Comparison of Q Learning applied to the Full MDP and to the Endogenous MDP for the traffic network problem ($N=200, T=400$).}
	\label{fig:traffic}
\end{figure}
Figure~\ref{fig:traffic} confirms that both Endo-Q learners are able to learn much faster than the Q-learner that is given the full MDP reward and that they are able to match the performance of the oracle.

\subsection{Problem 2: Linear system with 2-d exogenous state}\label{Sec:2D_toy_independent}

Let $X_{1,t}$ and $X_{2,t}$ be the exogenous state variables and $E_t$ be the endogenous state. The state transition function is defined as
\begin{equation}
\begin{split}
&X_{1,t+1} = 0.9 X_{1,t} + \epsilon_1,\\
&X_{2,t+1} = 0.7 X_{2,t} + \epsilon_2,\\
&E_{t+1} = 0.4 E_{t} + A_{t} + 0.1 X_{1,t}+ 0.1 X_{2,t} + \epsilon_3,
\end{split}
\end{equation}
where $\epsilon_1 \sim \mathcal{N}(0,0.16)$ and $\epsilon_2 \sim \mathcal{N}(0,0.04)$ and $\epsilon_3 \sim \mathcal{N}(0,0.04)$.

The observed state vector $S_t$ is a linear mixture of the hidden exogenous and endogenous states $S_t = M [X_{1,t}, X_{2,t}, E_{t}]^{\top}$,where
\[M=\begin{bmatrix}
&0.3,&0.6,& 0.7\\
&0.3,&-0.7,& 0.2\\
&0.6,& 0.3,& 0.2
\end{bmatrix}
\]
\begin{figure}
\centering
	\includegraphics[width=3.5in]{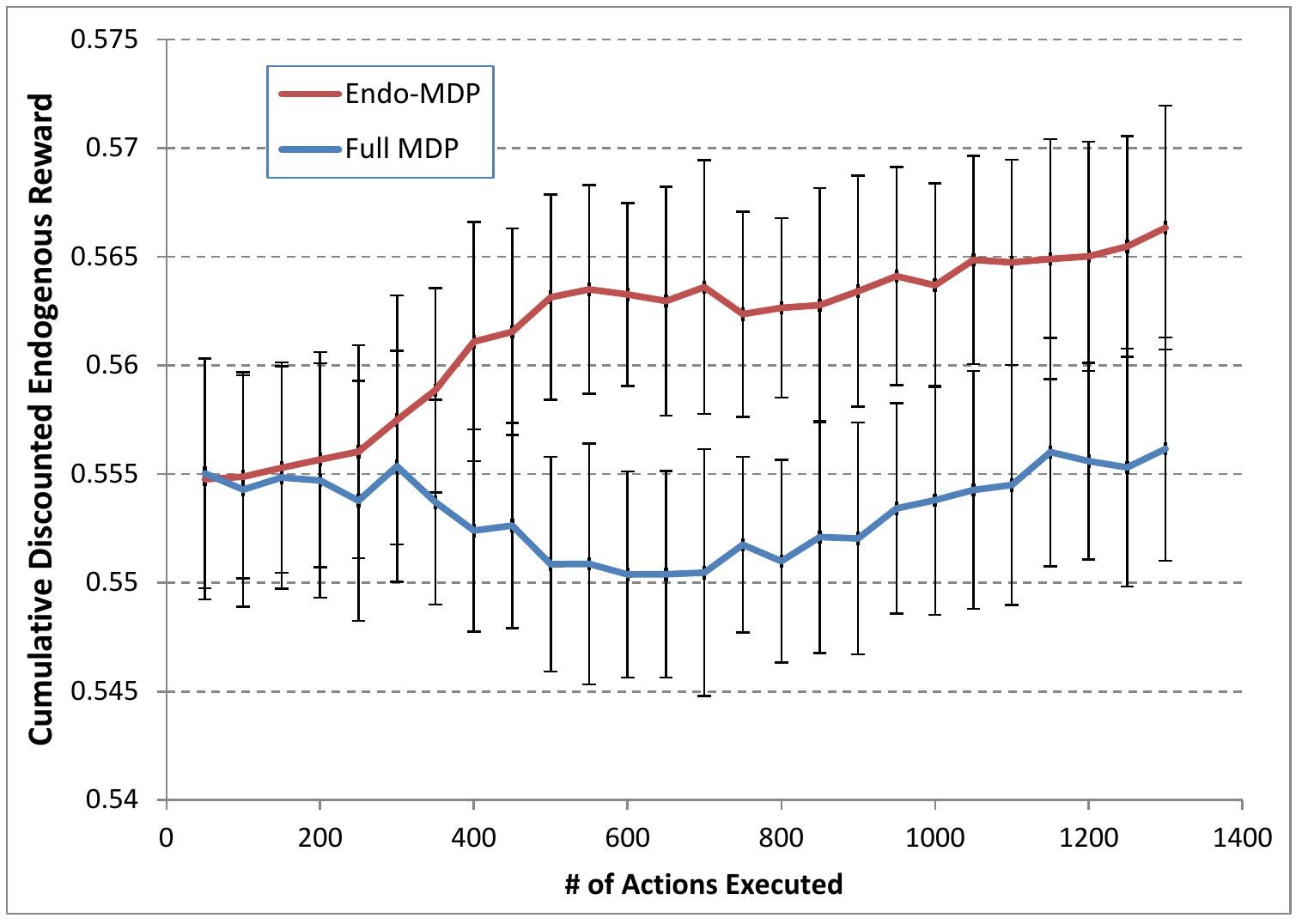}
	\caption{Comparison of Q Learning applied to the Full MDP and to the Endogenous MDP for a 3-d linear system with two coupled exogenous dimensions}
	\label{fig:toy2}
\end{figure}

\begin{figure}
\centering
	\includegraphics[width=3.5in]{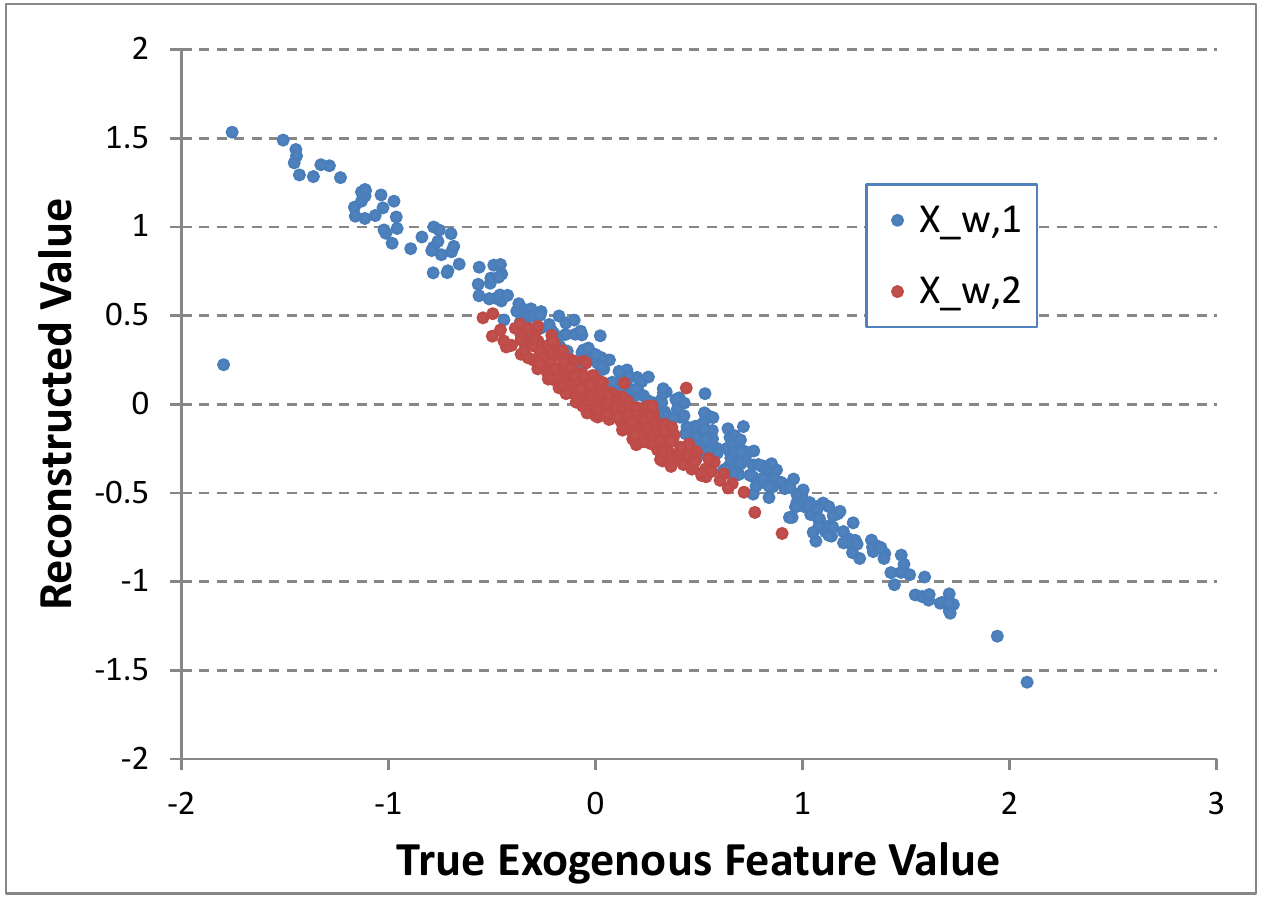}
	\caption{Comparison of Q Learning applied to the Full MDP and to the Endogenous MDP for a 3-d linear system with two coupled exogenous dimensions}
	\label{fig:toy2-reconstruction}
\end{figure}

The reward at time $t$ is defined as $R_t = R_{x,t} + R_{e,t}$, where $R_{x,t}$ is the exogenous reward $R_{x,t} = - X_{1,t} - X_{2,t}$
and $R_{e,t}$ is the endogenous reward $R_{e,t} = = \exp[|E_t - 3|/4]$. Figure~\ref{fig:toy2-reconstruction} shows that with the exception of a few extreme states, the learned $W_x$ successfully reconstructs the values of $X_1$ and $X_2$. 

\subsection{Problem 3: 5-d linear system with 3-d exogenous state}\
Let $X_{1,t}$, $X_{2,t}$, $X_{3,t}$ be the exogenous state variables and $E_{1,t}$, $E_{2,t}$ be the endogenous state variables. The state transition function is defined as:
\begin{equation*}
\begin{split}
&X_{1,t+1} = \frac{3}{5} X_{1,t} + \frac{9}{50} X_{2,t} + \frac{3}{10} X_{3,t} + \epsilon_1,\\
&X_{2,t+1} = {7}/{15} X_{2,t} + \frac{7}{50} X_{3,t} + \frac{7}{30} X_{1,t} + \epsilon_2,\\
&X_{3,t+1} = \frac{8}{15} X_{3,t} + \frac{8}{50} X_{1,t} + \frac{7}{30} X_{2,t} + \epsilon_3,\\
&E_{1,t+1} = \frac{13}{20} E_{1,t} + \frac{13}{40} E_{2,t} + A_{t} + 0.1 X_{1,t} + 0.1 X_{2,t} + \epsilon_4,\\
&E_{2,t+1} = \frac{13}{20} E_{2,t} + \frac{13}{40} E_{1,t} + A_{t} + 0.1 X_{2,t} + 0.1 X_{3,t} + \epsilon_5,
\end{split}
\end{equation*}
where $\epsilon_1 \sim \mathcal{N}(0,0.16)$, $\epsilon_2 \sim \mathcal{N}(0,0.04)$, $\epsilon_3 \sim \mathcal{N}(0,0.09)$, $\epsilon_4 \sim \mathcal{N}(0,0.04)$, and $\epsilon_5 \sim \mathcal{N}(0,0.04)$.

\begin{figure}
\centering
	\includegraphics[width=\columnwidth]{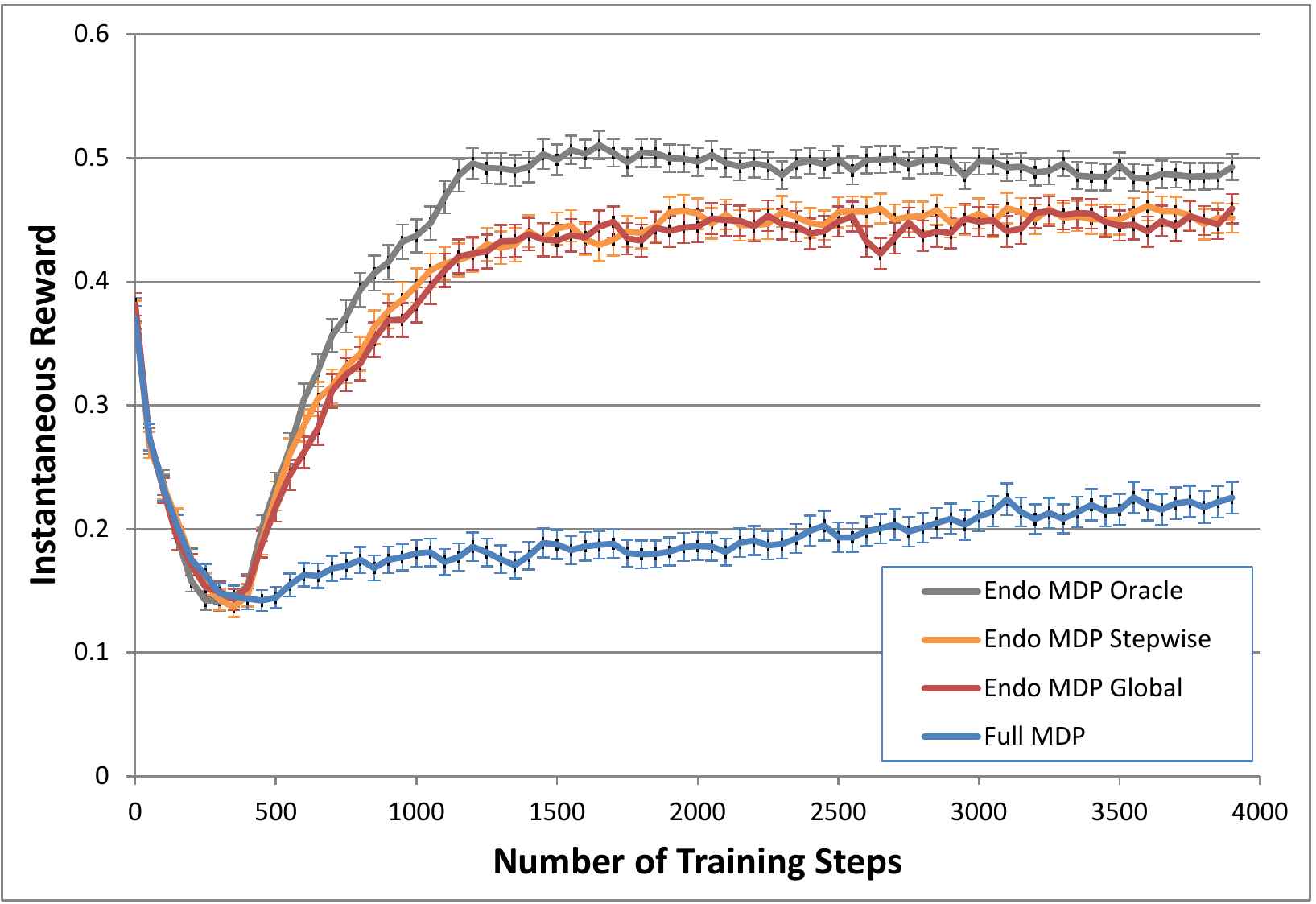}
	\caption{Comparison of Q Learning applied to the Full MDP and to the Endogenous MDP for a 5-d linear system with three coupled exogenous dimensions ($T=50$, $N=1000$).}
	\label{fig:toy-5-3}
\end{figure}

The observed state vector $S_t$ is a linear mixture of the hidden exogenous and endogenous states:
\[S_t = \begin{bmatrix}
0.3&0.3& 0.6&0.2&-0.4\\
0.6&-0.7& 0.3&0.5&-0.3\\
0.7&0.2& 0.2&-0.8&0.6\\
0.4&-0.2&-0.1&-0.2&0.9\\
0.9& 0.3&-0.2&0.7&-0.2
\end{bmatrix}\cdot\begin{bmatrix}
X_{3,t}\\
X_{2,t}\\
X_{1,t}\\
E_{2,t}\\
E_{1,t}
\end{bmatrix}.\]
The reward at time $t$ is defined as $R_t = R_{x,t} + R_{e,t}$, where $R_{x,t}$ is the exogenous reward $R_{x,t} = -1.4X_{1,t}-1.7X_{2,t}-1.8X_{3,t}$
and $R_{e,t}$ is the endogenous reward $R_{e,t} =\exp[-\frac{|E_{1,t} +1.5E_{2.t} - 1|}{5}]$. The action $A_t$ can take the discrete values $\{-1.0,-0.9,\dots,0.9,1.0\}$.

The PCC threshold was set to $0.1$ for this problem.


Figure~\ref{fig:toy-5-3} shows the performance of four Q Learning algorithms: ``endo oracle'' is trained on the true endogenous reward, ``endo stepwise'' and ``endo global'' are trained on the estimated endogenous reward after applying either the stepwise or the global optimization methods to estimate $W_x$, and ``full MDP'' is trained on the original MDP. 
Q learning on ``full MDP'' is very slow, whereas both ``endo stepwise'' and ``endo global'' are able to learn nearly as quickly as ``endo oracle''. There is no apparent difference between the two optimization methods. 

\end{document}